# Key, Value, Compress: A Systematic Exploration of KV Cache Compression Techniques

Neusha Javidnia[1], Bita Darvish Rouhani[2], Farinaz Koushanfar[1]

[1]University of California San Diego, USA
[2]NVIDIA, USA

## 1  Abstract

Large language models (LLMs) have demonstrated exceptional capabilities in generating text, images, and video content. However, as context length grows, the computational cost of attention increases quadratically with the number of tokens, presenting significant efficiency challenges. This paper presents an analysis of various Key-Value (KV) cache compression strategies, offering a comprehensive taxonomy that categorizes these methods by their underlying principles and implementation techniques. Furthermore, we evaluate their impact on performance and inference latency, providing critical insights into their effectiveness. Our findings highlight the trade-offs involved in KV cache compression and its influence on handling long-context scenarios, paving the way for more efficient LLM implementations.

## 2  Introduction

Large language models (LLMs) have rapidly advanced in recent years, utilizing the transformer structure across diverse architectures, including encoder-only, decoder-only, and encoder-decoder models. Encoder-only models, such as BERT [1], excel at understanding and extracting context from text, making them ideal for tasks like sentence classification and named entity recognition. Encoder-decoder models, such as T5 [2], integrate both encoding and decoding capabilities, achieving high performance in tasks that require both comprehension and generation, including translation and summarization. Decoder-only models, like GPT [3], specialize in generating coherent and contextually relevant text, making them powerful tools for applications such as text generation and dialogue systems.

Among these architectures, decoder-based models particularly benefit from Key-Value (KV) caching due to their autoregressive nature, where tokens are generated sequentially with each step conditioned on the entire preceding context. KV caching facilitates this process by storing intermediate representations of previously processed tokens, thereby eliminating the need to recompute these representations at each decoding step. This approach significantly enhances computational efficiency and reduces memory requirements. While encoder-based models typically process input tokens in parallel and thus do not necessitate KV caching, decoder-based models can leverage KV caching during the decoding phase in generation tasks. This is especially advantageous for tasks involving long input sequences or multimodal models, as it reduces redundant computations.

Emerging applications of LLMs increasingly require long-context inputs containing thousands of tokens or more, such as in retrieval-augmented generation (RAG) [4], document summarization, in-context learning, accumulated conversation histories, and analyzing domain-specific knowledge texts. As the number of tokens increases, demands on the key-value (KV) cache and memory rise significantly, impacting both storage and computation costs. For instance, a 65B parameter model with grouped-query attention [5] and 8-bit KV quantization requires around 86GB of GPU memory to handle 512K tokens — exceeding the capacity of a single H100-80GB GPU [6].

A number of recent methods have been proposed to compress the KV cache and reduce its memory requirements. Broadly, such approaches can be categorized based on compression applied across layers, heads, tokens, or the hidden dimension. Compression across layers involves sharing KV weights across multiple layers, as seen in recent variants of T5, where self-attention and cross-attention mechanisms share the same KV weights [7]. Compression across heads involves sharing KV weights among single or grouped heads, leading to multi-query [12] or grouped-query attention [5], as incorporated in LLaMA 3 models [8]. Compression across tokens includes techniques like token pruning and summarization. Mamba state-space-based models [9] which eliminate the need for a KV cache entirely, fall within this category. Finally, compression across hidden dimensions utilizes quantization to introduce structured sparsity, reducing memory requirements while retaining essential information.

In this paper, we provide a comprehensive categorization of KV cache methods, systematically organizing them based on their distinct compression strategies. We delve into the various categories of KV cache compression, analyzing their underlying principles and implementation techniques. These implementations include training the model from scratch, requiring post-training adjustments, or operating in a training-free manner without the need for additional fine-tuning. Additionally, we assess the impact of these methods on model accuracy and computational latency. Through this detailed comparison, we highlight the strengths, limitations, and trade-offs of each approach, offering insights into their overall effectiveness.

## 3  Background

### 3.1  Attention Mechanism

The attention mechanism is a powerful approach in modern neural networks, allowing models to focus on specific parts of the input sequence based on their relevance to the current output. Given an input sequence represented by a matrix X, attention computes a set of queries Q, keys K, and values V as:

$$Q = XW_Q, \quad K = XW_K, \quad V = XW_V$$

where $W_Q$, $W_K$, and $W_V$ are learned weight matrices. The attention score between each query and key is calculated using a dot product, followed by a softmax operation for normalization:

$$\text{Attention}(Q, K, V) = \text{softmax}\left(\frac{QK^T}{\sqrt{d_k}}\right) V$$

where $d_k$ is the dimensionality of the key vectors, scaling the dot product to avoid large values that could destabilize training. The resulting attention scores are then used as weights in a weighted sum of the values V, allowing the model to focus on the most relevant parts of the input sequence.

### 3.2  Key-Value (KV) Cache

In large language models (LLMs), token generation occurs sequentially, with each token generated one at a time. To optimize efficiency, the model reuses previously computed key-value pairs stored in the KV cache, which holds the keys K and values V produced at each layer for past tokens. This caching allows the model to access these precomputed values in later steps without needing to recompute them.

Formally, at each time step t, the attention mechanism uses the cached K and V pairs from previous tokens along with the current query $Q_t$ as follows:

$$\text{Attention}(Q_t, [K_1, K_2, \ldots, K_{t-1}], [V_1, V_2, \ldots, V_{t-1}])$$

The KV cache is updated through two key stages: (i) the *prefill* stage, in which the input sequence is processed to initialize the KV cache for each transformer layer, and (ii) the *decoding* stage, where the KV cache is incrementally updated as tokens are generated in an autoregressive manner. This approach allows the model to efficiently attend to previous tokens, minimizing computational load and conserving memory.

The size of the KV cache linearly increases with the number of prompt tokens, which can lead to significant memory overhead, especially for lengthy prompts or when processing multiple prompts in parallel. As each layer in the transformer model holds its own set of key-value pairs, the total memory requirement grows not only with the sequence length but also with the model depth, resulting in scalability challenges.

## 4 KV Cache Compression Techniques

In this section, we categorize KV cache compression methods and outline their structure. These approaches are broadly classified based on the level at which compression is applied, including layers, attention heads, tokens, and hidden dimensions. In the following subsections, we delve into the specifics of KV cache compression across layers, attention heads, and tokens.

For compression along the hidden dimensions, techniques such as quantization are employed. Quantization reduces the precision of the stored key and value vectors, typically by converting high-precision floating-point representations into lower-precision formats [10]. This method effectively decreases memory usage and computational overhead while maintaining sufficient accuracy for inference tasks. As quantization methods are well-established and do not require structural modifications to the model, they are discussed here as a general approach rather than in a dedicated subsection.

KV cache compression is particularly effective for tasks involving extensive input contexts, as opposed to generation-heavy tasks with more output than input. It maximizes cache throughput and reduce the overall memory footprint, thereby optimizing GPU utilization and improving scalability for handling large datasets. These KV cache compression techniques can be implemented at various stages, including during training, post-training, or even without retraining. While some methods may affect model accuracy—either positively or negatively—all are designed to enhance inference speed, aiming to balance performance improvements with resource efficiency.

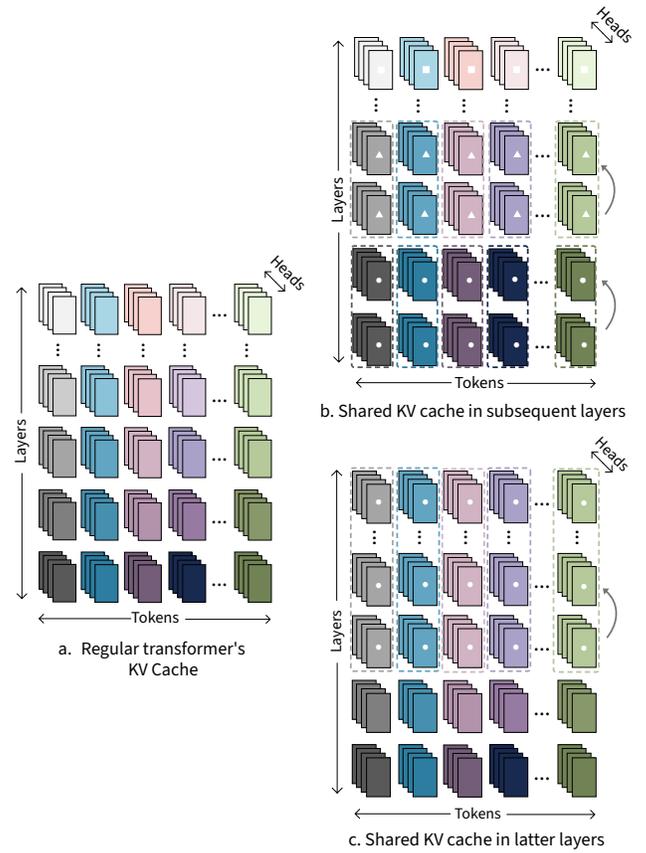

Figure 1: Illustration of KV Cache with *Cross-Layer Attention*. The KV cache is shared across layers, with two variations depicted: in (b) the KV cache is shared among subsequent layers, while in (c), it is shared only among the latter layers. This shared KV cache mechanism reduces redundant storage and computational overhead while allowing layers to leverage previously computed key-value pairs.

### 4.1 Compression across layers

Layer-based KV cache compression techniques can be broadly categorized into *Cross-Layer Attention* and *Layer-Selective Attention*.

#### 4.1.1 Cross-Layer Attention

In *Cross-Layer Attention*, as shown in Figure 1, KV weights are shared across multiple layers. This approach allows a unified set of KV weights to be applied to a group of layers, reducing the need to compute distinct KV weights for each layer. By reusing KV representations across designated layers, this method effectively reduces both memory usage and computational load. For instance, the KV cache can be shared among consecutive layers or only in the latter layers.

An example of Cross-Layer Attention is provided by *You Only Cache Once (YOCO)* [6], where only half of the decoder layers perform self-attention while the remaining layers perform cross-attention. In this design, KV cache values from the final self-attention module are shared across subsequent decoder layers, significantly reducing memory usage in the latter portions of the model. Implemented in a language model trained from scratch, YOCO achieves superior performance compared to Transformers across various model sizes and training token scales, as evidenced by experimental results.

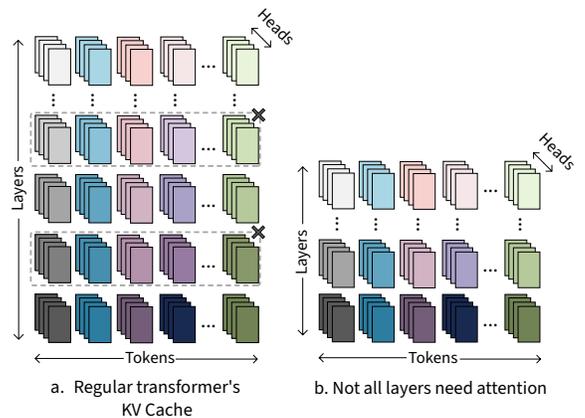

Figure 2: Illustration of KV Cache with *Layer-Selective Attention*. This is based on the principle that not all layers require the self-attention mechanism. By selectively pruning specific attention layers, computational complexity and memory usage can be significantly reduced while maintaining performance. In (a), specific layers are pruned, leading to a compressed KV cache, as illustrated in (b).

#### 4.1.2 Layer-Selective Attention

Conversely, *Layer-Selective Attention*, as illustrated in Figure 2, is based on the principle that not all layers require attention mechanism. By selectively pruning specific attention blocks, this technique minimizes computational and memory demands while preserving performance.

An example of Layer-Selective Attention is the *Attention Drop* approach proposed in [11]. This method introduces a joint layer-drop strategy that leverages cosine similarity between module inputs and outputs to identify redundancy. Attention layers deemed redundant are selectively pruned during the post-training stage on pretrained models. For example, in LLaMA-3-70B, half of the attention layers were pruned while achieving comparable performance to the original model.

### 4.2 Compression Across Heads

Three examples of head-wise KV cache compression are *Multi-Query Attention (MQA)*, *Grouped-Query Attention (GQA)*, and *Multi-Head Latent Attention (MLA)*, which are explained below.

The term *Multi-Query Attention (MQA)* was first introduced by Shazeer in [12], presenting a variation of the traditional multi-head attention mechanism. In this approach, the keys and values are shared across different attention heads, significantly reducing memory bandwidth requirements during incremental decoding.

*Grouped-Query Attention (GQA)*, as proposed by Ainslie et al. [5], generalizes Multi-Query Attention by dividing query heads into G groups, with each group sharing a single key and value head. When G = 1, GQA is equivalent to MQA, where all heads are combined into a single group sharing one key/value head. Conversely, when G = H, GQA becomes equivalent to Multi-Head Attention (MHA), with each head maintaining its own key/value pair. A comparison of the KV cache structure for GQA and MQA is shown in Figure 3 (b) and (a), respectively. GQA achieves performance comparable to its base model, with a slight drop, while outperforming MQA. This grouping is achieved through a post-training process, where the model initializes by mean-pooling the key and value projection matrices from all heads into the specified number of grouped heads. This flexible design balances memory efficiency and attention granularity based on the choice of G.

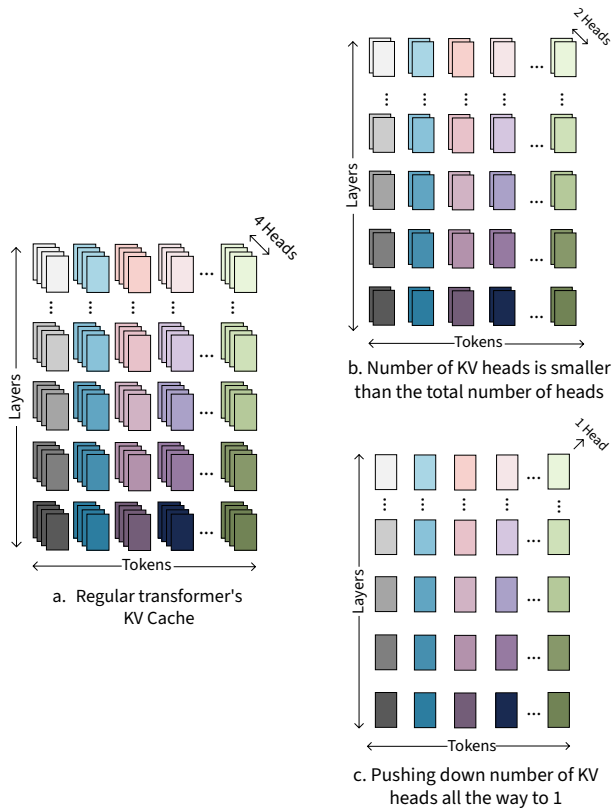

Figure 3. *Multi-Query Attention (MQA)* and *Grouped-Query Attention (GQA)* KV cache structures. (a) illustrates a standard transformer with four heads. (b) depicts GQA, where query heads are grouped into two, reducing the number of key-value pairs stored in the KV cache while preserving diverse attention patterns. (c) represents MQA, where all query heads share a single set of key-value pairs, significantly lowering memory requirements at the expense of reduced head diversity.

*Multi-Head Latent Attention (MLA)* was first introduced in DeepSeek-V2 [13] and was later utilized in DeepSeek-V3 [14]. MLA employs low-rank joint compression on attention keys and values, reducing the Key-Value (KV) cache size during inference. As shown in Figure 4(b), the mechanism compresses the KV cache into a latent representation using a down-projection matrix and later reconstructs it via up-projection matrices for keys and values. These projection matrices are trainable, enabling efficient memory usage while preserving critical attention information.

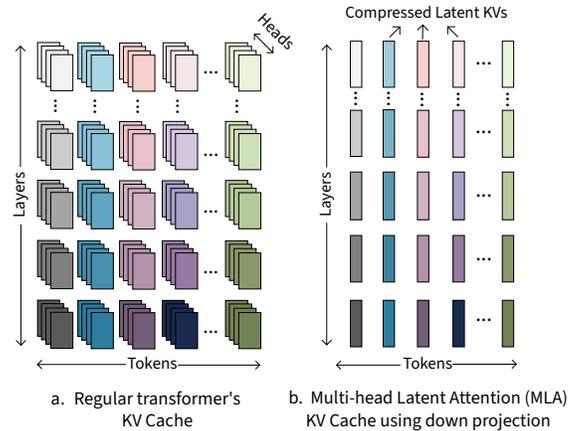

Figure 4: Comparison of the KV cache in a standard transformer and *Multi-Head Latent Attention (MLA)*, introduced in DeepSeek-V2 [13]. MLA compresses KV representations into a latent space using a down-projection matrix for efficient storage, then reconstructs them during computation.

### 4.3 Compression across tokens

This type of compression includes several research directions. The first category is *Structured State Space Models (SSMs)* [16, 17] and their variants, such as *Mamba* [9], which eliminate the need for a KV cache. The second involves pruning the attention matrix, with the goal of zeroing out as many entries as possible to reduce computational load. The third approach focuses on compressing the length of the input context to optimize memory usage. [15]. For the latter two approaches, we highlight a few examples, though numerous other methods have been explored extensively in NLP research.

State Space Models (SSMs) and their variants, such as Mamba, offer significant computational and memory efficiency. Unlike transformer-based architectures, which rely on explicit key-value (KV) caching for autoregressive decoding, Mamba leverages a discretized state-space representation that maintains a recurrent hidden state, allowing for efficient long-sequence modeling. The memory structure of SSMs is illustrated in Figure 5(b). In this figure, $l$ represents the number of layers in the model, and $h_t^l$ denotes the hidden state at a specific time step.

Mamba replaces the quadratic complexity of regular attention with an O(N) scaling by iteratively updating a compact hidden state rather than storing and attending to all previous tokens. During training, Mamba scales linearly with sequence length, a key advantage over regular attention mechanisms, which require $O(N^2)$ operations. During inference, Mamba computes each new output based on the previous state, eliminating the need to recompute the entire sequence history or cache previous elements.

However, while Mamba models offer significant efficiency gains, they do not always match transformers in accuracy across all tasks.

Transformers excel in tasks requiring complex, fine-grained attention patterns, whereas Mamba is particularly effective for applications where long-range dependencies are critical.

An advanced variant, the *Mamba2-Hybrid* architecture [18], addresses some of these limitations by improving in-context learning and information recall. To ensure higher accuracy for such tasks, Mamba2-Hybrid integrates Mamba-2 layers, self-attention layers, and multi-layer perceptron (MLP) layers, evenly distributed throughout the network. This hybrid design combines the computational efficiency of Mamba with the contextual and memory capabilities of self-attention, achieving superior performance compared to standard Mamba models in tasks requiring context understanding and memory. The memory structure of this model is illustrated in Figure 5(c).

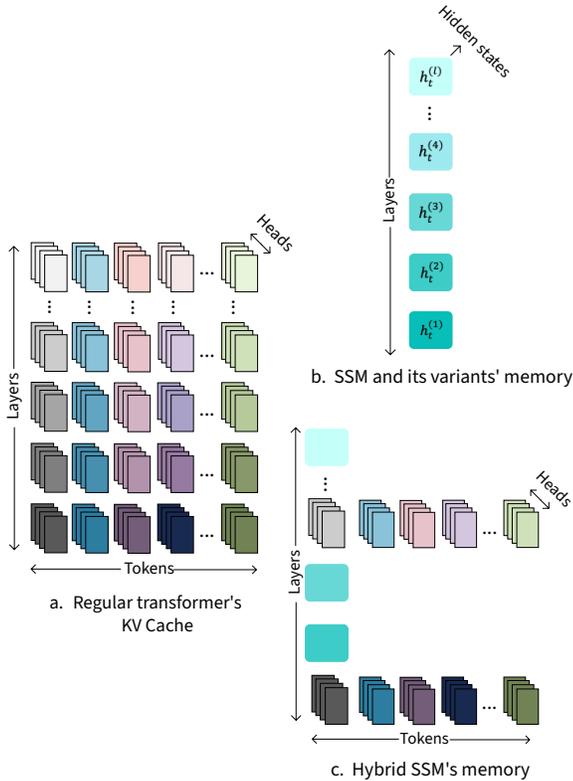

Figure 5. Comparison of computational memory in SSM-based models (b, c) and the Transformer's KV cache mechanism (a). Unlike Transformers, which rely on a KV cache (a) to store and retrieve past key-value pairs, SSMs (b) maintain a recurrent hidden state, eliminating the need to store or recompute the entire sequence history. This enables efficient long-sequence modeling. Since SSM-based models do not require a KV cache, we illustrate their computational memory. (c) depicts Mamba2-Hybrid, a hybrid architecture that integrates Mamba-2 layers, self-attention layers, and MLP layers. By combining these components, Mamba2-Hybrid addresses key limitations of pure SSM models, enhancing in-context learning and improving information recall.

The second category, illustrated in Figure 6(b), focuses on pruning tokens based on the attention map to enhance efficiency. For example, *LazyLLM* [19] performs layer-wise token pruning at each generation step, leveraging the attention map to retain only the most relevant tokens. Tokens retained in later layers form a subset of those used in earlier layers, enabling progressive pruning. This approach integrates with language models to accelerate generation without requiring fine-tuning, achieving negligible accuracy drops across multiple tasks. Additionally, *FastGen* [20] employs model profiling based on prompt encoding outcomes to determine the optimal compression strategy for each attention head. For instance, it evicts long-range contexts in attention heads focused on local information and discards non-special tokens in heads that prioritize special tokens.

Fine-tuned using open-sourced instruction-tuning datasets, FastGen delivers efficient compression with minimal loss in generation quality. This approach compresses the input context without requiring additional training of the language model, achieving efficiency gains with performance below that of baseline models.

The third category emphasizes compressing long-context inputs using either another language models or specialized algorithms to effectively reduce context length, as illustrated in Figure 6(b). For instance, *Context-aware Prompt Compression (CPC)* [22] applies sentence-level compression by removing less relevant sentences from the prompt using a context-aware sentence encoder. *Token Compression Retrieval Augmented (TCRA) LLM* [23] uses a fine-tuned T5-based model to either summarize text or perform semantic compression by selectively eliminating low-impact words. *LLM-Lingua* [24], in contrast, leverages a small language model to assess each prompt token's perplexity, removing tokens with lower perplexity values. Building on this, *LongLLMLingua* [25] employs a question-aware, coarse-to-fine compression strategy by first estimating document-level question relevance and then applying LLMLingua. In this coarse-grained approach, the perplexity of the question, conditioned on different contexts, guides the selection of essential content. All these methods operate in a training-free manner, focusing solely on compressing the input context. Notably, CPC and LongLLMLingua demonstrate improved accuracy, likely due to a regularization effect, while TCRA achieves near-baseline performance with minimal deviation.

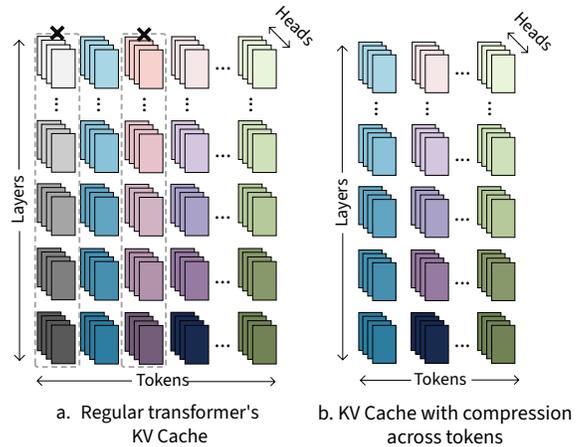

Figure 6. Illustration of KV Cache with token pruning and input compression techniques. These methods include pruning tokens or sentences or compressing the input context length using another language model, aiming to reduce computational load.

*Dynamic Memory Compression (DMC)* [26] is another technique designed to optimize the memory usage of LLMs during inference. At each time step, DMC predicts a decision variable that determines whether to append the current key and value representations to the cache or to merge them with the most recent entry through a weighted average. This method retrofits existing LLMs and, in some cases, even improves their performance, likely due to the additional fine-tuning steps involved.

## 5   Evaluation

In this section, we present the evaluation results of the discussed methods, focusing only on those based on standard benchmarks and LLMs as reported in their original papers. These evaluations span diverse tasks and datasets, which, as discussed further below, are not directly comparable. Our primary emphasis is on the relative performance of each method to facilitate a comparative analysis. Additionally, we include latency evaluations, acknowledging that these were conducted under varying conditions and may not be directly comparable. This comprehensive overview aims to provide insights into the strengths and limitations of each approach.

Table 1: Performance comparison of models trained from scratch. The Baseline/Model Average represents the average performance across the dataset based on the specified metric(s), and Relative Accuracy indicates the percentage increase or decrease in performance.

| Model | Compression Dim. | Baseline | Baseline Avg. | Model Avg. | Relative Accuracy |
|---|---|---|---|---|---|
| YOCO-3B* | Layers+Heads | OpenLLaMA-3B-v2 | 61.9 | 63.4 | 2.42% |
| *Dataset: ARC-C, ARC-E, BoolQ, HellaSwag, OBQA, PIQA, Winogrande, SciQ; Metric: LM Eval Harness | | | | | |
| Mamba-2.8B* | Tokens | RWKV-3B | 59.6 | 63.3 | 6.21% |
| *Dataset: LAMBADA, WinoGrande, PIQA, ARC-E, HellaSwag, ARC-C; Metric: Accuracy, Accuracy Normalized | | | | | |
| Mamba2-8B* | Tokens | GPT3-8B | 62.35 | 64.16 | 2.90% |
| *Dataset: WinoGrande, PIQA, ARC-E, HellaSwag, ARC-C, MMLU; Metric: LM Eval Harness (Accuracy, Accuracy Normalized) | | | | | |
| Mamba2-Hybrid-8B* | Tokens | GPT3-8B | 53.17 | 55.82 | 4.98% |
| *Dataset: WG, PIQA, HellaSwag, ARC-E, ARC-C, MMLU, OpenBook, TruthfulQA, PubMed, RACE, NQ, SquadV2; Metric: LM Eval Harness (Accuracy, Exact Match, F1) | | | | | |

Table 2: Performance comparison of models requiring post-training. The Baseline/Model Average represents the average performance across the dataset based on the specified metric(s), and Relative Accuracy indicates the percentage increase or decrease in performance.

| Model | Compression Dim. | Baseline | Baseline Avg. | Model Avg. | Relative Accuracy |
|---|---|---|---|---|---|
| Att. Drop-4* | Layers | Llama-2-13B | 68.2 | 68.5 | 0.44% |
| *Dataset: ARC-C, BoolQ, HellaSwag, MMLU, OBQA, PIQA, RTE, WinoGrande; Metric: LM Eval Harness (Accuracy, Accuracy (Norm), Exact Match) | | | | | |
| MQA-XXL* | Heads | T5-XXL | 47.2 | 46.6 | -1.27% |
| *Dataset: CNN, arXiv, PubMed, MediaSum, MultiNews, WMT, TriviaQA; Metric: R1, BLEU, F1 | | | | | |
| GQA-8-XXL* | Heads | T5-XXL | 47.2 | 47.1 | -0.21% |
| *Dataset: CNN, arXiv, PubMed, MediaSum, MultiNews, WMT, TriviaQA; Metric: R1, BLEU, F1 | | | | | |
| DMC 2x* | Tokens | Llama2-7B | 43.03 | 43.73 | 1.63% |
| *Dataset: MMLU, CS-QA, HumanEval; Metric: Accuracy, Pass@1 | | | | | |
| DeepSeek-V3 | Heads | Llama3.1-405B | 87.96 | 90.07 | 2.4% |
| *Dataset: ARC-C, ARC-E, HellaSwag, PIQA, Winogrande, MMLU, CMATH; Metric: Exact Match | | | | | |

### 5.1 Datasets and Metrics

The models are evaluated on a diverse set of datasets, including ARC-C and ARC-E [27], BoolQ [28], Hellaswag [29], OBQA [30], PIQA [31], WinoGrande [32], SciQ [33], MMLU [34], RTE [35], and TriviaQA [36]. Additional datasets such as CNN [37], arXiv [38], PubMed [39], MediaSum [40], MultiNews [41], WMT, LAMBADA [42], LongBench [43] (covering SingleDoc, MultiDoc, Summarization, FewShot, Synthesis, and Code), CS-QA [44], HumanEval [46], and CMath [47] are also included. Each method is evaluated on a subset of these datasets, with details of the datasets used for each method outlined in Tables 1, 2, and 3.

Models are evaluated using a variety of metrics tailored to specific tasks. Many methods leverage the LM Eval Harness [48], a comprehensive evaluation framework that includes metrics such as Accuracy, Normalized Accuracy, Exact Match, and F1 Score. For text generation tasks, commonly used metrics include ROUGE-1 (R1), BLEU, and F1 Score, especially for summarization and translation contexts. In addition, ROUGE-L, F1, Accuracy, and Edit Similarity (Edit Sim) are utilized to evaluate text similarity and structural alignment. For code generation and completion tasks, Pass@1 is used to measure success rates on the first attempt.

### 5.2 Methods and Baselines

Each method's evaluation is sourced from its original publication, with models and their respective baselines assessed across various parameter sizes. Performance comparisons are provided in Tables 1, 2, and 3, with latency speedup details highlighted in Table 4.

For methods involving training large language models (LLMs) from scratch, YOCO [6], Mamba, and their variants are evaluated. The YOCO model utilizes layer-wise compression combined with GQA [5] for head-wise compression, using OpenLLaMA-3B-v2 [49] as its baseline. For the Mamba-based models, Mamba [9] with 2.8B parameters is compared to RWKV-3B [50], while Mamba2 [51] and Mamba2-Hybrid [18], both with 8B parameters, are evaluated against GPT3-8B [52].

In approaches requiring post-training on pre-trained models, Attention Drop [11], MQA, GQA [5], DMC [26], and DeepSeek-V3 [14] are evaluated. The Attention Drop approach demonstrates results with a 4-attention-layer reduction from Llama2-13B [53]. MQA and GQA, configured with 8 groups, represent post-trained versions of T5-XXL [54] with 11B parameters and are compared to the original model. Similarly, DMC is applied to Llama2-7B [53], with evaluations conducted relative to this baseline. Finally, DeepSeek-V3 is compared against Llama3.1-405B [8].

Conversely, methods that can be applied directly without additional training include CPC [22], LazyLLM [19], LongLLMLingua [25], and H$_2$O [21]. All of these methods utilize token-based compression. CPC and LongLLMLingua are tested with GPT-3.5-turbo, while LazyLLM and H$_2$O are evaluated with Llama2-7B [53].

Finally, the hardware configurations for testing each method, including GPUs (H100, A100, V100), are detailed in Table 4. These setups outline the computational resources used for benchmarking each approach.

### 5.3 Performance Comparison

As shown in Tables 1, 2, and 3, which present models requiring training, post-training, and no training, respectively, KV cache compression strategies exhibit varying impacts on performance. These methods may enhance accuracy, leave it relatively unchanged, or result in accuracy drops.

Table 3: Performance comparison of models with a training-free approach. The Baseline/Model Average represents the average performance across the dataset based on the specified metric(s), and Relative Accuracy indicates the percentage increase or decrease in performance.

| Model | Compression Dim. | Baseline | Baseline Avg. | Model Avg. | Relative Accuracy |
|---|---|---|---|---|---|
| CPC* | Tokens | GPT3.5-turbo | 44 | 50 | 13.64% |
| *Dataset: LongBench; Metric: ROUGE-L, F1, Accuracy, Edit Sim | | | | | |
| LongLLMLingua* | Tokens | GPT3.5-turbo | 44 | 48.3 | 9.77% |
| *Dataset: LongBench (SingleDoc, MultiDoc, Summ., FewShot, Synth., Code); Metric: ROUGE-L, F1, Accuracy, Edit Sim | | | | | |
| LazyLLM* | Tokens | Llama2-7B | 32.65 | 32.29 | -1.10% |
| *Dataset: LongBench; Metric: ROUGE-L, F1, Accuracy, Edit Sim | | | | | |
| $H_2O$* | Tokens | Llama2-7B | 43.03 | 40.83 | -5.11% |
| *Dataset: MMLU, CS-QA, HumanEval; Metric: Accuracy, Pass@1 | | | | | |

Table 4: Comparison of model speedup across sequence lengths and baseline models. Sequence lengths are specified as *context + generation*. Latency speedup indicates the improvement achieved by each KV-cache compression method relative to its baseline on the specified hardware.

| Model | Sequence Length | Baseline | Latency Speedup | Type of Latency | Hardware |
|---|---|---|---|---|---|
| YOCO-3B | 32k[1] | Transformer[2] | 2.87x | Prefilling | H100-80GB GPUs |
| LLMDrop-Attn-4 | 512 to 4096[3] | Llama-2-13B | 1.05x | End to end | a A100-80GB GPU |
| CPC | 3376+68 | GPT-3.5-turbo - LongLLMLingua [25] | 10.93x | End to end | a A100-80GB GPU |
| LazyLLM | 3376+68 | Llama 2 7B | 2.03x, 1.33x | TTFT, End to End | A100 GPUs |
| DMC 8x | 2k + 2k | Llama 2 7B | 1.25x | Next token generation | a H100-80GB SXM |
| LongLLMLingua | 3376+68 | GPT-3.5-Turbo | 2.6x | End to end | a V100-32GB GPU |
| $H_2O$ | 2048+2048 | OPT-6.7B-FlexGen [56] | 1.86x | End to end | a A100-80GB GPU |
| FastGen | 4096+4096 | Llama 1-7B - DeepSeed [57] | 1.56x | End to end | 8 V100 GPUs |

[1] For the first row, sequence length refers to context length only.
[2] Baseline Transformer employs GQA [5], Flash-Decoding [55], and kernel fusion.
[3] For the second row, sequence length includes both context length and generated response length.

Generally, methods involving training models from scratch using KV cache compression strategies demonstrate improved accuracy relative to their baselines, albeit with the significant overhead of the costly LLM training process. For instance, Mamba-2.8B achieved a relative accuracy gain of 6.2% compared to its baseline model, as depicted in Table 1.

Conversely, methods requiring post-training, as shown in Table 2, tend to maintain relatively stable performance. For example, the Attention Drop method, applied to Llama-2-13B, exhibited a modest relative performance increase of 0.44%.

Finally, as depicted in Table 3, methods that do not require training and focus on input context compression also demonstrate notable improvements. For instance, the CPC and LongLLMLingua models achieved relative accuracy gains of 13.64% and 9.77%, respectively, compared to GPT-3.5-turbo. These results highlight the benefits of context reduction, showing how removing redundant information can help language models make more accurate predictions while providing a regularization effect.

## 5.4 Speedup Comparison

Table 4 presents the speedup of various methods compared with respect to their sequence lengths and baseline models. The results are sourced from original papers, and a common baseline is needed for more accurate comparison. Overall, all methods demonstrate a speedup relative to their base models. However, the extent of this speedup depends on factors such as hardware configurations, baseline models, sequence length, and batch size, which collectively influence inference time.

The sequence length represents the input context length and generated token length, providing a complete view of the workload. The table also highlights different latency types: 1) *Prefilling* time refers to the time required to process the input and prepare the Key-Value (KV) cache, excluding token generation. 2) *End-to-End* time captures the total latency, from input to the completion of the output sequence. 3) *Time-to-First-Token (TTFT)* measures the time to generate the first token, reflecting responsiveness. 4) *Next-Token Generation* represents the latency for generating each token after the first.

This comparison highlights the extent to which each method reduces computational requirements. This reduction can be achieved by methods such as pruning or summarizing tokens to decrease context length or by reducing computations across attention heads.

## 6 Future Work

This study conducted comparisons across various baselines, datasets, training configurations, and different hardware setups. While these comparisons provide valuable insights, a rigorous "apple-to-apple" evaluation remains necessary to ensure a fair and comprehensive assessment. Such an evaluation would involve using consistent baselines, datasets, batch sizes, and hardware configurations across all methods. We propose this as an important direction for future work.

Moreover, a significant gap exists in evaluating GPU memory usage and throughput (measured in tokens per second). This gap can be addressed by systematically comparing strategies that incorporate Key-Value (KV) cache compression under standardized conditions. Such research would provide valuable insights into the trade-offs between performance, memory efficiency, and computational throughput.

## 7 Conclusion

In this paper, we have systematically categorized and surveyed various Key-Value (KV) cache compression strategies, focusing on the compression across all dimensions of the KV cache. These strategies are instrumental in reducing memory footprint, optimizing GPU utilization, and enhancing both latency and throughput. We analyzed the performance of various KV cache compression methods relative to their respective baselines and datasets, highlighting their impact on model efficiency. Notably, KV cache compression proves particularly beneficial for tasks involving long-context inputs, as it effectively reduces redundancy and improves processing efficiency. While our study provides a comprehensive overview, a holistic evaluation of these methods remains an avenue for future research.